# Empirical Normalization for Quadratic Discriminant Analysis and Classifying Cancer Subtypes

Mark A. Kon
Dept. of Mathematics and Statistics
Program in Bioinformatics
Boston University, Boston, MA 02215
mkon@bu.edu

Nikolay Nikolaev
Dept. of Mathematics and Statistics
Boston University, Boston, MA 02215
niki@math.bu.edu

*Abstract*— We introduce a new discriminant analysis method (Empirical Discriminant Analysis or EDA) for binary classification in machine learning. Given a dataset of feature vectors, this method defines an empirical feature map transforming the training and test data into new data with components having Gaussian empirical distributions. This map is an empirical version of the Gaussian copula used in probability and mathematical finance. The purpose is to form a feature mapped dataset as close as possible to Gaussian, after which standard quadratic discriminants can be used for classification. We discuss this method in general, and apply it to some datasets in computational biology.

## I. INTRODUCTION

In high dimensional machine learning, formation of discriminants (e.g. linear discriminants in the case of support vector machines) is a common method for forming classifiers [1, 2]. A classic example of discriminant analysis occurs in parametric statistics, in which the quadratic Fisher discriminant is used to separate two classes defined on the same feature space which are assumed to obey Gaussian distributions [1, 3].

In theoretical probability, mathematical finance and statistics there are methods for the formation of standardized uniform and Gaussian distributions from arbitrary joint distributions, known as the method of copulas [4, 5]. Given a random vector $\mathbf{X} = (X_1, \ldots, X_d)$, the formation of uniform variables from the components of $\mathbf{X}$ is done by composing them with their own distribution functions $F_{X_i}(x)$ (assuming these are fully known). The further formation of Gaussian components is accomplished by composing each of these uniform components with the inverse univariate Gaussian distribution function $\phi^{-1}(\cdot)$. We will denote such a normalization map which maps a non-normal $\mathbf{X}$ into a $\mathbf{Z}$ with exactly normal components as $\mathbf{Z} = \Psi(\mathbf{X})$, and call $\Psi$ a *normalizing map*. Note that such a normalization is different from standard normalization methods in which only the mean and variance (and not the full distribution) are standardized.

In this paper we will study such maps $\Psi$ as feature maps on datasets $D$ for purposes of classification. For any continuous random vector $\mathbf{X}$, the above normalizing map $\Psi$ exists and is unique. For a given dataset $D = \{\mathbf{x}_i\}_{i=1}^n$ (viewed as a random sample from a single random variable $\mathbf{X}$) we can try to construct an empirical approximation to $\Psi$. Given such an approximating $\Psi$ (which we denote as an *empirical normalizing map*), it follows that using this as a feature map yields a dataset $\Psi(D) = \{\mathbf{z}_i = \Psi(\mathbf{x}_i)\}_{i=1}^n$ of feature vectors with normal components. Given two such datasets $D$ and $D'$ (representing two classes to be discriminated), the two corresponding empirical normalizing feature maps will be called $\Psi$ and $\Psi'$ (note that primes will not represent derivatives here). These maps can be used to normalize the two datasets as above in order derive the (now normal) joint densities of the transformed feature vectors. Using these densities, likelihood ratios can be applied to novel test vectors $\mathbf{x}$ in order to determine which class they belong to.

Thus our copula-based empirical feature maps $\Psi$ and $\Psi'$ will respectively map the case and control training datasets $D = \{\mathbf{x}_i\}_{i=1}^n$ and $D' = \{\mathbf{x}_i'\}_{i=1}^n$ of feature vectors into datasets $\Psi(D)$ and $\Psi'(D')$ with manifestly normal (empirical) components. These empirical feature maps are then fixed and applied in the same way to a test data point $\mathbf{x}$ in order to obtain likelihoods $L(\mathbf{x})$ and $L'(\mathbf{x})$ based respectively on the normal empirical distributions of the transformed case and control data. There is finally a likelihood ratio discriminant function $h(\mathbf{x}) = \dfrac{L(\mathbf{x})}{L'(\mathbf{x})}$ to perform the case/control classification on $\mathbf{x}$.

We will show here that an algorithm implementing the empirical feature maps $\Psi$ and $\Psi'$ from case and control datasets $D$ and $D'$ is simple, and based on the rankings of a component $x_i$ of a novel test feature vector $\mathbf{x}$ within the list of the $i^{th}$ components of the two training sets, first $D$ and then $D'$. To this extent, this algorithm allows the transformation of ranking scores into likelihood scores.

**Applications to simulated and real data.** We will apply this algorithm to some simulated classification problems, and two cancer classification datasets in bioinformatics.

Developing gene expression-based biomarkers for prediction of tissue types (e.g., various cancer phenotypes such as metastatic vs. non-metastatic [6, 7]) or response to therapy has been important since the start of gene expression technology development [8, 9]. Conventional linear discriminant analysis (LDA) and quadratic discriminant analysis (QDA) for binary classification often assume that features have approximately multivariate normal distributions [1, 3]. Such methods are therefore often suboptimal when dealing with non-normal and in particular heavy-tailed distributions [10]. The distributions of gene expression data are often significantly non-normal, and there is a large array of data normalization methods in use [11]. One normalization practice involves applying power transformations to data in the hope of achieving normality, though these are not always effective [12].

The empirical normalization feature map $\Psi$ transforms non-normally distributed feature vectors $\mathbf{x} = (x_1,\ldots,x_n)$ into feature vectors $\Psi(\mathbf{x}) = \mathbf{z} = (z_1,\ldots,z_n)$ with manifestly normal marginal distributions of the $z_i$, which are then further assumed to have *jointly* normal multivariate distributions. As mentioned, the key for binary classification is to perform empirical normalization separately on case and on control training data, using separate normalization feature maps $\Psi'$ and $\Psi$. For a test vector $\mathbf{x}$, the values of the two multivariate normal density functions are divided to form a discriminant in the form of a likelihood ratio, which determines the assigned class. Though such normalizations have been done on genomic data [13, 14], they have not (to our knowledge) been used as feature maps for classification.

**Advantages and drawbacks.** An advantage of empirical normalization discriminant analysis is that it can be invariant under monotonic transformations of features $x_i$ (and to that extent is again a rank-based method). It works well with non-normally distributed data, is easy to implement, and forms a straightforward rank-based algorithm for computing likelihood ratios. However, a drawback is the current necessity of prior feature selection to reduce dimensionality, since estimation is troublesome when the number of features is too high compared to the sample size. More specifically, instabilities in estimation of individual marginal densities for too many variables accumulate in such situations and make numerical values unreliable.

As is shown in the simulations below, the performance of this method is comparable to standard quadratic discriminant analysis (QDA) on data that initially have jointly normal distributions. However, it is significantly better for heavy-tailed feature distributions, as will be shown for joint densities given by the multivariate *t*-distribution. Experiments with some benchmark bioinformatics datasets have resulted in overall better performance of the empirical discriminant method when compared with QDA.

## II. CLASSIFICATION BASED ON EMPIRICAL DENSITY ESTIMATION AND NORMALIZATION

Consider training data containing two sets: cases $D = \{\mathbf{x}_i\}_{i=1}^{n}$ and controls $D' = \{\mathbf{x}_i'\}_{i=1}^{n'}$, assumed to be selected independently from two underlying probability distributions $\mu$ (for cases) and $\mu'$ (for controls) in a $d$-dimensional feature space $\mathcal{X}$.

**Construction of the feature map.** We first construct the feature map $\Psi$ used to map the case dataset $D$. The feature map $\Psi: \mathcal{X} \to \mathcal{Z}$ will map into a new feature space $\mathcal{Z}$ of the same dimensionality $d$ as $\mathcal{X}$, and be defined by the following property. We require that the empirical distribution $\hat{\mu} = \sum_{i=1}^{n} \delta_{\mathbf{x}_i}$ of the dataset $D$ be mapped by composition with $\Psi$ into an approximate multivariate normal empirical distribution $\hat{\mu} \circ \Psi^{-1}$ (defining $\hat{\mu} \circ \Psi^{-1}(A) = \hat{\mu}(\Psi^{-1}(A))$). This requirement is enforced only on the marginal distributions, making this method valid only to the extent that the mapped multivariate empirical distribution $\hat{\mu} \circ \Psi^{-1}$ is approximately jointly normal.

Writing sample vectors $\mathbf{x} \in D$ as $\mathbf{x} = (x_1,\ldots,x_d)$ (and similarly for vectors $\mathbf{x}' \in D'$), we will define $\Psi$ as follows. Let $\mathbf{X} = (X_1,\ldots,X_d)$ be a random vector with the empirical distribution $\hat{\mu}$ of $D$, i.e., $\mathbf{X}$ is a random element $\mathbf{x}_i \in D$. We define the first component of our feature map on $D$ as the function
$$\mathbf{F}(\mathbf{X}) = \mathbf{F}_\mathbf{X}(\mathbf{X}) = \mathbf{F}_\mathbf{X}(X_1,\ldots,X_d) = (F_{X_1}(X_1),\ldots,F_{X_d}(X_d)).$$
Here for a random variable $X$, we define $F_X(x) = \mathbb{P}(X \leq x)$ to be its distribution function. For the discrete empirical random variable $X_i$ with $n$ values, the variable $F_X(X)$ is a discrete approximation to the uniform distribution, being a sum $\frac{1}{n}\sum_{i=1}^{n} \delta_{i/n}$ of equally spaced point masses $\delta_{i/n}$ at points $\{i/n : i = 1,\ldots,n\}$. Thus each component of
$$F(\mathbf{X}) = (Y_1,\ldots,Y_d)$$
Has exactly the above discretized uniform distribution. The further formation of normal random variables is done by applying the inverse distribution function $\Phi^{-1}$ of the standard multivariate normal. That is,
$$\Phi^{-1}(\mathbf{Y}) = \Phi^{-1}(Y_1,\ldots,Y_d) = (\phi^{-1}(Y_1),\ldots,\phi^{-1}(Y_d))$$
$$= (Z_1,\ldots,Z_d) = \mathbf{Z}, \qquad (1)$$
Where $\phi(x)$ denotes the cumulative distribution function of the standard Gaussian with density
$$g(x) = \frac{d\phi}{dx}(x) = \frac{1}{\sqrt{(2\pi)}} e^{-x^2/2}. \qquad (2)$$
The result is a variable $\mathbf{Z} = \Phi^{-1}(\mathbf{F}_\mathbf{X}(\mathbf{X}))$ whose marginal densities are standard normal distributions, discretized into $n$ point masses with equal normal probabilities between them. We define empirical covariance matrix $\Sigma = \text{Cov}(\mathbf{Z})$ by $\Sigma_{ij} = \text{cov}(Z_i, Z_j)$.

As an example, if $\mathbf{X}$ represents a biological gene expression array, with $X_j$ the measured expression of gene $j$, then $\mathbf{Z}$ will represent the normalization of $\mathbf{X}$, with discretized normal components $Z_i$ (in the above sense). To the extent the resulting $\mathbf{Z}$ is *jointly* normal, i.e., the components have a joint density which is a discretization of the density function

$$\rho_{\mathbf{Z}}(z_1,\ldots,z_d) = G(\mathbf{z}) = (2\pi)^{-d/2} e^{-\frac{1}{2}\mathbf{z}^T \Sigma^{-1} \mathbf{z}} \quad (3)$$

Then the density (3) will also be an exact probabilistic representation of the *joint* dependences of the components $Z_j$ of the mapped feature vector $\mathbf{Z} = \Psi(\mathbf{X}) = \Phi^{-1}(\mathbf{F}_\mathbf{X}(\mathbf{X}))$. Thus $\Psi$ defines our normalization feature map.

We note that $\Psi$ acts component wise on coordinates, being a composition of the functions $\Phi^{-1}$ and $F_\mathbf{X}$ which also act componentwise. Thus we can write $\Psi = (\psi_1,\ldots,\psi_d)$ for the case dataset $D$ with an analogous function $\Psi'$ defined for the control data. From above, we have $\psi_i(x_i) = \phi^{-1}(F_{X_i}(x_i))$.

**Balancing the empirical distributions.** For the purpose of balancing the empirical distribution of the feature mapped dataset $\Psi(D)$, we modify the above definition to be $\psi_i(x) = \phi^{-1}\left(F_{X_i}(x) - 1/(2n)\right)$. Effectively this moves each point mass in the uniform distribution to the left by an amount $1/(2n)$. This balances the resulting empirical uniform distribution about $x = 1/2$, and the final normal distribution about $z = 0$. More specifically, (defining $x_{ij}$ as the $j^{th}$ component of data point $\mathbf{x}_i$), the additional term $1/(2n)$ still leaves (for fixed $j$) the uniformized points $\{F_{X_i}(x_{ij})\}_{i=1}^n$ separated from each other by $1/n$, while the extreme points (for $i=1$ and $i=n$) are separated by $1/(2n)$ from 0 and 1, respectively. Note also that $\psi_i(x_{ij}) = \phi^{-1}(F_{X_i}(x_{ij}))$ are division points of the same quantiles of the standard normal distribution as the points $F_{X_i}(x_{ij})$ are of the uniform distribution. Thus the standard normal area between two successive points in the sequence $\{\psi_i(x_{ij})\}_{i=1}^n$ is $1/n$, and the areas to the left and right of the extreme points ($i=1,n$ respectively) are $1/(2n)$.

**Forming the discriminant.** Following the above definitions for dataset $D$, corresponding primed quantities are similarly defined for dataset $D'$, in particular with the analogous feature map $\Psi'$ normalizing $D'$.

After feature maps $\Psi$ (for case data $D$) and $\Psi'$ (for control data $D'$) are defined, the marginal distributions of the random variables $\mathbf{Z} = \Psi(\mathbf{X})$ and $\mathbf{Z}' = \Psi'(\mathbf{X}')$ are manifestly (discretized) normals, and the feature-mapped datasets $\Psi(D) = \{\mathbf{z}_i\}_{i=1}^n$ and $\Psi'(D') = \{\mathbf{z}_i'\}_{i=1}^{n'}$ have (discretized) standard normal empirical marginal distributions.

The empirical covariance matrices $\Sigma = \text{Cov}(\mathbf{Z})$ and $\Sigma' = \text{Cov}(\mathbf{Z}')$ of the $\mathbf{z}$-data from $D$ and $D'$, respectively, are then approximations to covariances of the true multivariate normal distributions of the case and control data (similarly feature mapped to have normal marginals).

Now we form the discriminant for classifying a test sample $\mathbf{x} = (x_1,\ldots,x_d)$ as case or control. If we knew the true underlying densities $L(\mathbf{x})$ and $L'(\mathbf{x})$ of the case and the control random variables $\mathbf{X}$ and $\mathbf{X}'$, discrimination would be straightforward. The assigned class of the test point $\mathbf{x}$ would be determined by the discriminant likelihood ratio $h(\mathbf{x}) = \dfrac{L(\mathbf{x})}{L'(\mathbf{x})}$ with a classification as a case point if $h > 1$ and a control if $h < 1$. Here we form an estimate of the same discriminant $h$ based on the normalized data $\Psi(\mathbf{x})$ and $\Psi'(\mathbf{x})$.

**Formation of Gaussian random variables.** For a test feature vector $\mathbf{x}$, the likelihood (true density) $L(\mathbf{x})$ (under the assumption that $\mathbf{x}$ is in the case group) takes the form

$$L(\mathbf{x}) = \left|\frac{\partial \mathbf{z}}{\partial \mathbf{x}}\right| G(\mathbf{z}(\mathbf{x})), \quad (4)$$

which is the backward mapping of the normalized density $G(\mathbf{z}) = (2\pi)^{-d/2} (\det \Sigma)^{-1/2} e^{-\frac{1}{2}\mathbf{z}^T \Sigma^{-1} \mathbf{z}}$ in $\mathbf{z}$ into the original variable $\mathbf{x}$, using the feature map $\mathbf{z} = \Psi(\mathbf{x})$.

Here $\left|\dfrac{\partial \mathbf{z}}{\partial \mathbf{x}}\right|$ is the Jacobian of $\Psi(\mathbf{x})$, which (since $\Psi$ acts only component-wise) can be written as the product $\left|\dfrac{\partial \mathbf{z}}{\partial \mathbf{x}}\right| = \dfrac{\partial z_1}{\partial x_1} \cdots \dfrac{\partial z_d}{\partial x_d}$. Since the Jacobian requires existence of a derivative of $\Psi$, we use the underlying (non-empirical) $\Psi$ in this formula (defined as above in terms of the underlying non-empirical random variables $X_i$, assumed to be continuous), and then discuss the empirical estimation of these derivatives.

It can be shown that $\dfrac{\partial z_i}{\partial x_i}$ is an (inverted) ratio of the single variable (non-empirical) densities, i.e., $\dfrac{\partial z_i}{\partial x_i} = \dfrac{\rho_{X_i}(x_i)}{g(z_i)}$. Here again we are defining as $\rho_{X_i}(x)$ the density of the underlying (non-empirical) $X_i$, and $g(z)$ is the (standard normal) density of $z$ given in (1).

Thus we can estimate the underlying density function of the case data at the test point $\mathbf{x}$ to be:

$$L(\mathbf{x}) = \left|\frac{\partial \mathbf{z}}{\partial \mathbf{x}}\right| G(\mathbf{z}(\mathbf{x})) = \left|\frac{\partial z_1}{\partial x_1} \cdots \frac{\partial z_d}{\partial x_d}\right| (2\pi)^{-d/2} (\det \Sigma)^{-1/2} e^{-\mathbf{z}^T \Sigma^{-1} \mathbf{z}/2}$$

$$= \frac{\rho_{X_1}(x_1)\ldots\rho_{X_d}(x_d)}{g(z_1)\ldots g(z_d)} (2\pi)^{-d/2} (\det \Sigma)^{-1/2} e^{-\mathbf{z}^T \Sigma^{-1} \mathbf{z}/2}$$

$$= \rho_{X_1}(x_1)\ldots\rho_{X_d}(x_d)(\det \Sigma)^{-1/2} e^{-\mathbf{z}^T (\Sigma^{-1}-1) \mathbf{z}/2}.$$

This gives the likelihood ratio as (all primed quantities refer to the control dataset $D'$)

$$h(\mathbf{x}) = \frac{L'(\mathbf{x})}{L(\mathbf{x})} = \frac{\rho'_{X_1'}(x_1)\ldots\rho'_{X_d'}(x_d)(\det \Sigma')^{-1/2} e^{-\mathbf{z}'^T (\Sigma'^{-1}-1) \mathbf{z}'/2}}{\rho_{X_1}(x_1)\ldots\rho_{X_d}(x_d)(\det \Sigma)^{-1/2} e^{-\mathbf{z}^T (\Sigma^{-1}-1) \mathbf{z}/2}}$$

$$= \frac{\rho'_{X_1'}(x_1)}{\rho_{X_1}(x_1)} \cdots \frac{\rho'_{X_d'}(x_d)}{\rho_{X_d}(x_d)} e^{\mathbf{z}^T(\Sigma^{-1}-1)\mathbf{z} - \mathbf{z}'^T(\Sigma'^{-1}-1)\mathbf{z}'/2} \left(\frac{\det \Sigma}{\det \Sigma'}\right)^{1/2}$$

$$= J_1(x_1)\ldots J_d(x_d) e^{\mathbf{z}^T(\Sigma^{-1}-1)\mathbf{z} - \mathbf{z}'^T(\Sigma'^{-1}-1)\mathbf{z}'/2} \left(\frac{\det \Sigma}{\det \Sigma'}\right)^{1/2} \quad (5)$$

with $J_j(x_j) = \frac{\rho'_{X_j'}(x_j)}{\rho_{X_j}(x_j)}$ representing the direct Jacobian ratio for the $i^{th}$ coordinate between the case and the control data. That is, $J_j(x_j)$ represents the ratio of the marginal densities of the control and case data restricted to the $j^{th}$ coordinate only, at the current value $x_j$ in the test feature vector.

**Estimating the Jacobian.** To estimate the value of the Jacobian $J_j(x_j)$, an empirical estimate of this one dimensional ratio of densities can be obtained as follows. We first rank the values $\{x_{ij}\}_{i=1}^n$ of the $j^{th}$ coordinates in $D$ into list which we denote as $\{x_{i;j}\}_{i=1}^n$, with the (ranked) elements now increasing in $i$. For the given component $x_j$ in test vector $\mathbf{x} = (x_1,\ldots,x_d)$ we compute the empirical estimator

$$J_j(x_j) \approx \frac{x_{i+k-1;j} - x_{i-k;j}}{x'_{i+k-1;j} - x'_{i-k;j}}, \quad (6)$$

where $x_{i;j}$ denotes the smallest element of (the $j^{th}$ coordinate of) the control data set which is larger than $x_j$, with the analogous definition for $x'_{i;j}$ relative to $x_j$ for the case dataset. The numerator of (6) above represents the distance between the $k^{th}$ point to the right and the $k^{th}$ point to the left of $x_j$ among the (ranked) empirical values $\{x_{i;j}\}_{i=1}^n$ of only the marginal coordinate $X_j$ the case dataset $D$. The denominator is defined similarly for $D'$ with respect to the same starting value $x_j$. We note that the empirical normalization method can be sensitive to the value of $k$, i.e., the width of the neighborhood within which the Jacobian is estimated. Optimal values of $k$ can be easily found through cross-validation.

**Equivalence with density estimation.** This classification procedure can be considered equivalent to a density estimation-based method done separately for the control and for the case data, in sense of equation(4). That is, the inverse feature map $\Psi^{-1}(\mathbf{z})$ composed with the empirical Gaussian density $G(\mathbf{z})$ of the normalized feature vector $\mathbf{z} = \Psi(\mathbf{x})$ yields the inferred density function $L(\mathbf{x})$ with marginal distributions $\rho_{X_i}(x_i)$. We recall again that under a map $\Psi^{-1}(\mathbf{z}): \mathcal{Z} \to \mathcal{X}$ of a variable $\mathbf{z}$ in a base space $\mathcal{Z}$, a probability density $G(\mathbf{z})$ on $\mathcal{Z}$ is transformed into the density $L(\mathbf{x})$ on $\mathcal{X}$ defined by

$$L(\mathbf{x}) = G(\mathbf{z}(\mathbf{x}))\left|\frac{\partial \mathbf{z}}{\partial \mathbf{x}}\right|, \quad (7)$$

again with the last term representing the Jacobian of $\Psi$. Thus (7) is an imputed joint density of the case data vector $\mathbf{X}$ in the variable $\mathbf{x}$, based on the normalization feature map $\Psi(\mathbf{x})$. Thus the likelihood ratio is simply the ratio of the imputed densities for the case and control distributions.

**This is a ranking-based approach.** As suggested above, this approach can be viewed as a ranking-based method. This holds to the extent that for any given test data point $\mathbf{x} = (x_1,\ldots,x_d)$, we need only know the ranking of each test component $x_j$ among the $j^{th}$ features $\{x_{i;j}\}_{i=1}^n$ in the case data $D = \{\mathbf{x}_i, y_i\}_{j=1}^n$ and also in the control data $D' = \{\mathbf{x}'_i, y'_i\}_{i=1}^{n'}$. The $j^{th}$ component $z_j = \psi_j(x_j)$ of the feature mapping $\Psi$ can then be computed directly by converting (for each fixed $j$) this ranking into a standard Gaussian density quintile via the mapping $r(x) \to \phi^{-1}(r(x))$, where $r(x) = \frac{R(x)}{n}$, is the normalized ranking of $x$, and $\phi$ is the distribution function of the standard univariate Gaussian. Here $R(x)$ the ranking (with 1 representing smallest and $n$ representing largest) of the number $x_j$ in the set $D_j$ (of all $j^{th}$ components of vectors in $D$) which is closest to $x$ from above. Applied individually to all components of a feature vector $\mathbf{x}$, we denote the ranking transformation by $\mathbf{z} = \Phi^{-1}(\mathbf{r}(\mathbf{x}))$ ($\Phi$ as in (1) is the standard Gaussian distribution function in $d$ dimensions). Note that the empirical distribution of $\mathbf{z}$ is a (discretized) normal, with each component having a standard normal empirical distribution.

If we ignore the Jacobian factors (6) between the feature values $x_j$ and their normalized values $z_j$, then (for both the control dataset $D'$ and the case dataset $D$) this mapping for each component $x_j$ allows purely ranking-based classification of any new data point through the comparison of the values and $G(\Phi^{-1}(\mathbf{r}(\mathbf{x})))$ $G'(\Phi^{-1}(\mathbf{r}'(\mathbf{x})))$, which are the Gaussian empirical density functions of the rank-normalized case data sets, $\Phi^{-1}(\mathbf{r}(D))$ with $D$ replaced by $D'$ for the control data. Here again

$$G(\mathbf{z}) = (2\pi)^{-d/2}(\det \Sigma)^{-1/2} e^{-\mathbf{z}^T\Sigma^{-1}\mathbf{z}/2}, \quad (8)$$

where $\Sigma_{ij} = \text{cov}(Z_i, Z_j)$, with $Z_i$ the empirical random variable normalized from the empirical random variable $X_i$, here for the case dataset. Replacing $G$ and $\Sigma$ by their primed counterparts gives the formula for the case dataset $D'$.

This purely rank-based discriminant, defined as $\overline{h}(\mathbf{x}) = \frac{G'(\Phi^{-1}(\mathbf{r}'(\mathbf{x})))}{G(\Phi^{-1}(\mathbf{r}(\mathbf{x})))}$ is equivalent to, with omission of the ratio of Jacobian factors. It nevertheless provides a novel way to impute probabilities (likelihoods) to ranking-based comparisons for classification purposes, in particular when there is reason to believe that the covariance structures of the case and control datasets are not too different. The discriminant $\bar{h}(\mathbf{x})$ not only takes relative values of ranking vectors $\mathbf{r}(\mathbf{x})$ into account, but also incorporates the correlations of rankings through the Gaussian model(8). The formation of such rank-based discriminants will be discussed in more detail in future work.

## III. RESULTS

*A. Simulated Data.* As indicated earlier, the empirical normalization method can be used effectively in reduced data dimensions. For initially high dimensional datasets this can be accomplished after dimensional reduction. Here we apply this method to simulated datasets in two, ten, twenty and fifty dimensions, and to two biological datasets.

We simulated case and control data respectively by using pairs $\rho(\mathbf{x})$ and $\rho'(\mathbf{x})$ of underlying multivariate distributions with small tails (multivariate normal) and heavy tails (multivariate *t* distribution with 1 degree of freedom). The sample sizes used ranged from 100 to 10,000, with a 5 to 1 ratio in training to test data, with balanced cases and controls. The average accuracy of 100 simulations is computed. In the tables below, (T) denotes the (theoretical) Bayes optimal estimator, which classifies a test point $\mathbf{x}$ according as whether the (underlying) ratio $\rho(\mathbf{x})/\rho'(\mathbf{x})$ is greater or less than 1. Method (S) is a quadratic kernel SVM using the same training and test data. Method (Q) assumes $\rho$ and $\rho'$ to be Gaussian, estimates means and covariances on training data, and classifies test data with standard Gaussian quadratic discriminant analysis. Method (E) is the present empirical normalization method. Method (J) consists of the same empirical normalization method, with the replacement of the empirically determined Jacobian component (6) with its (known) theoretical value. Methods T and J are based on underlying distribution knowledge and are benchmarks for best theoretical performances respectively of all classifiers (for T), and our empirical method E (for J). In the the application (E) of empirical normalization, the rank-based discriminant was calculated as in above, with $\Sigma$ and $\Sigma'$ determined empirically by the feature mapped (empirically normalized) datasets $\Psi(D)$ and $\Psi'(D')$ for the simulated case and control data. For each coordinate $x_i$ in the test data vector $\mathbf{x}$, the corresponding Jacobian was estimated using (6), with a choice of $k$ (the number of neighbors to the left and right of $x_i$ in the training data sets used to estimate $J_i(x_i)$) selected for the Jacobian ratio to cover 80% of the dataset for normal, and 10% for the $t_{[df=1]}$. Results (all in terms of percentage classification accuracy) are in Tables I and II below.

|  | Dim | 2D | | | | | 10D | | | | |
|---|---|---|---|---|---|---|---|---|---|---|---|
|  |  | E | J | Q | S | T | E | J | Q | S | T |
| I | 100 | 71 | 72 | 75 | 70 | 76 | 69 | 69 | 73 | 65 | 83 |
| I | 1000 | 74 | 75 | 75 | 72 | 75 | 79 | 80 | 81 | 79 | 82 |
| I | 10000 | 75 | 75 | 75 | 71 | 75 | 81 | 82 | 82 | 81 | 82 |
| II | 100 | 76 | 76 | 78 | 73 | 78 | 83 | 83 | 85 | 75 | 92 |
| II | 1000 | 79 | 79 | 80 | 77 | 80 | 91 | 91 | 92 | 92 | 93 |
| II | 10000 | 79 | 79 | 79 | 77 | 79 | 92 | 92 | 93 | 92 | 93 |
| III | 100 | 82 | 83 | 84 | 82 | 85 | 91 | 91 | 94 | 81 | 98 |
| III | 1000 | 85 | 85 | 85 | 83 | 85 | 96 | 96 | 97 | 97 | 98 |
| III | 10000 | 85 | 85 | 85 | 83 | 85 | 97 | 97 | 98 | 98 | 98 |
| IV | 100 | 89 | 89 | 91 | 88 | 91 | 96 | 96 | 98 | 90 | 99 |
| IV | 1000 | 90 | 90 | 91 | 89 | 90 | 99 | 99 | 99 | 99 | 99 |
| IV | 10000 | 90 | 90 | 90 | 89 | 90 | 99 | 99 | 99 | 99 | 99 |

|  | Dim | 20D | | | | | 50D | | | | |
|---|---|---|---|---|---|---|---|---|---|---|---|
|  |  | E | J | Q | S | T | E | J | Q | S | T |
| I | 100 |  |  |  |  |  |  |  |  |  |  |
| I | 1000 | 84 | 84 | 87 | 82 | 89 | 83 | 83 | 90 | 81 | 97 |
| I | 10000 |  |  |  |  |  |  |  |  |  |  |
| II | 100 |  |  |  |  |  |  |  |  |  |  |
| II | 1000 | 92 | 92 | 95 | 92 | 96 | 90 | 90 | 97 | 86 | 99 |
| II | 10000 |  |  |  |  |  |  |  |  |  |  |
| III | 100 |  |  |  |  |  |  |  |  |  |  |
| III | 1000 | 96 | 96 | 99 | 97 | 99 | 96 | 95 | 99 | 92 | 100 |
| III | 10000 |  |  |  |  |  |  |  |  |  |  |
| IV | 100 |  |  |  |  |  |  |  |  |  |  |
| IV | 1000 | 99 | 99 | 100 | 99 | 100 | 99 | 99 | 100 | 95 | 100 |
| IV | 10000 |  |  |  |  |  |  |  |  |  |  |

Table I: Accuracies of five data classification methods in 2, 10, 20 and 50 dimensions with normal data on balanced datasets. Cases I-IV represent different degrees of separation for two normal distributions used to select equal numbers of simulated case and control data. Size represents total number of data points used – data were split 5 to 1 among training and test data. (E) – empirical normalization method; (J) – empirical normalization using theoretical underlying Jacobian; (Q) – quadratic discriminant method (assuming normality); (S) – support vector machine (SVM); (T) – optimal Bayes method based on knowledge of theoretical underlying densities, using optimal Bayes classifier. For all data the control distribution is centered at the origin. For 2-d data the four case distributions are centered along the $x_1$ axis at 1, 2, 3, and 4. Covariances are $\begin{pmatrix} 2 & 2 \\ 2 & 5 \end{pmatrix}$ for the control and $\begin{pmatrix} 10 & .5 \\ .5 & 3 \end{pmatrix}$ for all 4 cases. For 10, 20 and 50 dimensions the case centers are at 1, 2, 3 and 4 along $x_1$, and the off-diagonal covariances have form $\Sigma_{ij} = \rho^{|i-j|}$ with $\rho = .1$ for control and $\rho = .9$ for case in row I; $\rho = .6$ for control and $\rho = .4$ for case in row II; $\rho = .7$ for control and $\rho = .3$ for case in row III; $\rho = .8$ for control and $\rho = .2$ for case in row IV. Variances are $\rho_{ii} = i$ for control and $2i$ for case.

We observe that performances of all the methods in both tables typically plateau with data sets of size 100 to 1,000 (of which 5/6 are training data), and do not improve beyond these numbers of data points. Note the relative performance of the empirical normalization method as compared to others is related not to statistical error (variance). Rather it is based on the above-mentioned potential bias in assuming that the true density $\rho$, when empirically normalized to the density $G(\mathbf{z})$ in(8), is *jointly* as well as marginally normal.

|   | Dim | 2D | | | | | 10D | | | | |
|---|---|---|---|---|---|---|---|---|---|---|---|
|   |   | E | J | Q | S | T | E | J | Q | S | T |
| I | 100 | 57 | 64 | 54 | 53 | 73 | 53 | 56 | 52 | 56 | 64 |
|   | 1000 | 67 | 69 | 52 | 52 | 72 | 59 | 59 | 51 | 56 | 64 |
|   | 10000 | 67 | 68 | 51 | 52 | 72 | 59 | 59 | 50 | 56 | 64 |
| II | 100 | 65 | 67 | 54 | 59 | 74 | 68 | 70 | 57 | 76 | 79 |
|   | 1000 | 70 | 71 | 52 | 60 | 75 | 74 | 75 | 51 | 76 | 79 |
|   | 10000 | 71 | 72 | 51 | 58 | 76 | 75 | 76 | 50 | 76 | 79 |
| III | 100 | 71 | 75 | 56 | 66 | 79 | 75 | 76 | 58 | 80 | 84 |
|   | 1000 | 75 | 76 | 52 | 67 | 79 | 80 | 81 | 51 | 83 | 85 |
|   | 10000 | 76 | 76 | 51 | 66 | 80 | 81 | 82 | 50 | 83 | 85 |
| IV | 100 | 75 | 78 | 57 | 71 | 82 | 79 | 80 | 68 | 83 | 87 |
|   | 1000 | 79 | 80 | 52 | 72 | 83 | 84 | 85 | 52 | 87 | 89 |
|   | 10000 | 79 | 80 | 51 | 73 | 83 | 84 | 85 | 50 | 87 | 88 |

|   | Dim | 20D | | | | | 50D | | | | |
|---|---|---|---|---|---|---|---|---|---|---|---|
|   |   | E | J | Q | S | T | E | J | Q | S | T |
| I | 100 | | | | | | | | | | |
|   | 1000 | 51 | 57 | 50 | 57 | 65 | 50 | 58 | 54 | 59 | 67 |
|   | 10000 | | | | | | | | | | |
| II | 100 | | | | | | | | | | |
|   | 1000 | 54 | 73 | 51 | 78 | 81 | 55 | 79 | 55 | 85 | 88 |
|   | 10000 | | | | | | | | | | |
| III | 100 | | | | | | | | | | |
|   | 1000 | 58 | 80 | 53 | 84 | 86 | 53 | 68 | 55 | 79 | 83 |
|   | 10000 | | | | | | | | | | |
| IV | 100 | | | | | | | | | | |
|   | 1000 | 60 | 85 | 54 | 88 | 90 | 57 | 83 | 56 | 88 | 91 |
|   | 10000 | | | | | | | | | | |

Table II: Same methods for heavy-tailed multivariate $t_{[df=1]}$ distributions shifted with varying degrees of separation and covariance structures (I-IV). The data are generated through dividing multivariate normally distributed variates by scaled square root of independent $\chi^2_{[1]}$ random variables and then adding mean vectors. Means in 2, 10, 20 and 50 dimensions are the same as for Table I with the exception of the first case, where $x_1 = 1$ is replaced by $x_1 = .5$. Covariance structures for $t_{[1]}$ are generated similarly to Table I for normal: the multivariate normals from the numerator have the same off-diagonal covariance structure as in Table 1 and the ratio with the scaled $\chi^2_{[1]}$ is multiplied by the corresponding standard deviations.

The table entries show how the method compares under different types and separations for the case and control distributions, in 2,10, 20 and 50 dimensions. In particular, aside from statistical fluctuations, the J numerical accuracies should dominate the E accuracies, given that an underlying theoretical (known) Jacobian is used for J.

It is worth noting that the empirical normalization method performs similarly to classic Quadratic Discriminant Analysis when the data are normally distributed and significantly better in the heavy-tailed distribution case, except in 50 dimensions. Therefore, there is a clear advantage to normalizing data empirically, instead of assuming they are already normal. Another observation is that the performance of the E-method increases with increasing dimensionality of normal samples from 2 to 10 but then changes only slightly at 20 and 50 dimensions. In the $t_{[df=1]}$ case it is also higher at 10 than 2 dimensions but then decreases at 20 and 50 dimensions. A plausible explanation is that the assumption of joint normality is more likely to be violated in higher dimensions. We also note that the SVM performance steadily increases in the $t_{[df=1]}$ case with increasing dimensionality while, for the normal case, it improves with increasing dimensionality from 2 to 10 to 20; however it decreases when dealing with 50 features.

Based on the above observations we claim that the advantages of our method are mostly pronounced when classifying non-normal data and in small dimensions. The latter observation motivates employment of a procedure for feature selection before this algorithm is used.

*B. Experimental Data*

There are a number of current works on classification of cancer phenotypes featuring robust dimensionally reduced feature sets. One recent example is due to Bienkowska, *et al.* [15] using so-called convergent random forests (CRF) for feature selection and classification. Another method is due to Geman, Tan and collaborators [16-19]. Bienkowska, *et al.* [15] identified a set of eight genes which differentiated breast cancers which later metastasized from ones which did not metastasize [7]. We used a well-defined subset of seven of these gene expressions to predict metastasis in the same breast cancer dataset. In addition, the Singh, et al. [20] prostate dataset (cancer vs. normal) was studied by Bienkowska, et al., using 5 significant genes, again using convergent random forest. The percentage accuracies for these dimensionally reduced benchmark datasets ($d = 7, 5$ respectively) in [15], including our empirical normalization method (E) along with QDA and SVM are in the table below. Mimicking the reference paper, training and test data are randomly selected in the prostate cancer dataset but are predefined for the breast cancer dataset. For the prostate cancer dataset all methods have similar performances. However, with the breast cancer dataset the accuracy of the empirical normalization method dominates that of the CRF and even more so the SVM and QDA accuracies.

|   | QDA (Q) | CRF | SVM (S) | Emp (E) |
|---|---|---|---|---|
| **Prostate** | 94 | 95 | 93 | 94 |
| **Breast** | 79 | 84 | 68 | 89 |

## IV. Conclusion

Classification based on empirical density estimation can be problematic when the number of features exceeds the number of samples. This can be addressed by reducing numbers of parameters in density estimates, which is effectively what is done here using empirical normalization marginals. In $d$ dimensions, once marginal empirical normalizations are fixed, if the normalized data are jointly normal, then only empirical covariance parameters in the (case and control) covariance matrices $\Sigma$ and $\Sigma'$ must be determined from case and control data $D$ and $D'$.

In theory an empirical likelihood ratio of this type should be valid in arbitrary dimension $d$. Nevertheless, a ratio based on such effective density estimates can involve very small denominators and suffer instabilities for large $d$. Thus values of $d$ should ideally be kept under small via feature selection. One alternative to be investigated for stabilizing such results involves use of fixed additive constants ('pseudocounts') to increase small denominators.

A more serious limitation is related to the fact (as illustrated when the $t_{[df=1]}$ distribution in $\mathbb{R}^{10}$ above is normalized) that the constructed univariate normal distributions may not necessarily generate multivariate normals. There are goodness-of-fit statistical tests for multivariate normality of the feature mapped vectors $\mathbf{z} = \Psi(\mathbf{x})$, but such tests may be insufficiently sensitive and difficult to interpret. After empirically guaranteeing normal marginals, determining how a potential lack of joint normality in data vectors $\mathbf{z}$ can affect classification is an important topic for further investigation.

## V. Acknowledgment

This work was partially supported by NIH grant 1R21CA13582-01 and NIH grant 1R01GM080625-01A1.